# Compass-Embedding v4: Robust Contrastive Learning for Multilingual E-commerce Embeddings

Pakorn Ueareeworakul[1], Shuman Liu[1], Jinghao Feng[1], Ling Hu[1], Zhantang Shi[1], Chengqi Sun[1], Liang Yao[1], Panyi Ouyang[1], Haibo Zhang[1, *] and Anxiang Zeng[1, *]

[1]Shopee LLM Team, *Corresponding Author

As global e-commerce rapidly expands into emerging markets, the lack of high-quality semantic representations for low-resource languages has become a decisive bottleneck for retrieval, recommendation, and search systems. In this work, we present Compass-Embedding v4, a high-efficiency multilingual embedding framework specifically optimized for Southeast Asian (SEA) e-commerce scenarios, where data scarcity, noisy supervision, and strict production constraints jointly challenge representation learning. Compass-Embedding v4 addresses three core challenges. First, large-batch contrastive training under mixed task supervision introduces systematic false negatives that degrade semantic alignment. We propose Class-Aware Masking (CAM), a lightweight modification to the InfoNCE objective that suppresses invalid in-batch negatives and improves semantic discrimination without altering training efficiency. Second, low-resource SEA languages suffer from limited and uneven data coverage. We construct a diversified training corpus through context-grounded synthetic data generation, cross-lingual translation, and structured e-commerce data construction, enabling robust multilingual and domain-specific learning. Third, production deployment requires high-throughput inference while preserving embedding quality. We combine robustness-driven large-batch training with spherical model merging to mitigate catastrophic forgetting, and optimize inference via vLLM and FP8 quantization. Extensive evaluations across multilingual benchmarks and proprietary e-commerce tasks show that Compass-Embedding v4 achieves state-of-the-art performance on major SEA languages, significantly outperforming general-purpose embedding models in domain-specific retrieval and classification, while maintaining competitive performance on high-resource languages.

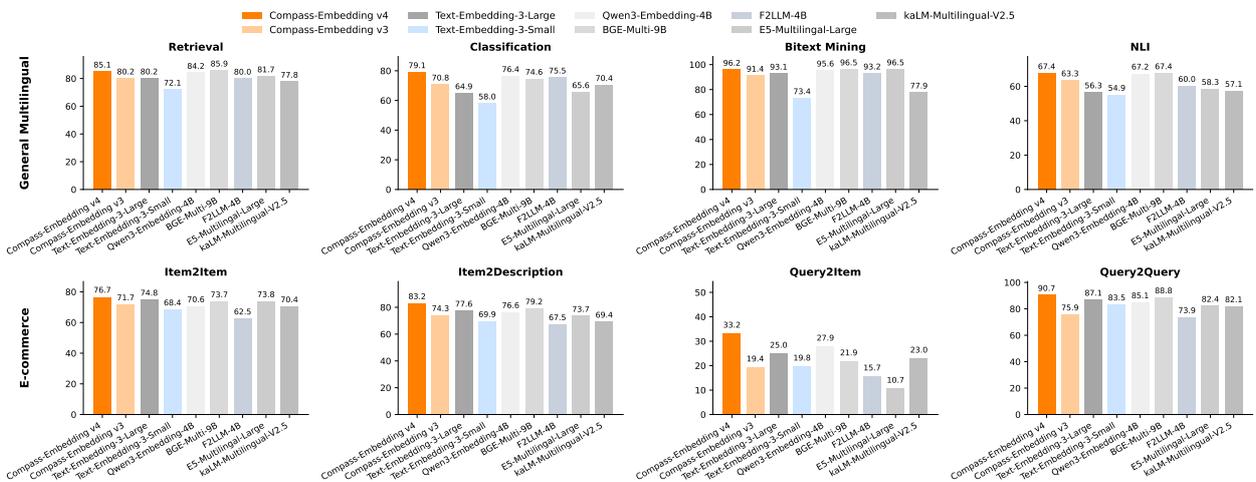

Figure 1 | Benchmark performance of Compass-Embedding v4.

# 1. Introduction

Text embedding models have emerged as a fundamental infrastructure component across modern natural language processing and information retrieval systems. By transforming discrete textual sequences into continuous dense vector representations, these models enable semantic similarity computation at scale, powering critical applications including semantic search, recommendation, ranking, question answering, and, more recently, retrieval-augmented generation (RAG) (Karpukhin et al., 2020; Lewis et al., 2021; Liu et al., 2024, 2025) systems that tightly integrate retrieval with large language models. As RAG frameworks continue to gain adoption in both research and industrial settings, the quality, robustness, and efficiency of embedding models have become increasingly critical to end-to-end system performance.

Despite rapid progress, the development of embedding models remains highly asymmetric across languages and domains. Existing state-of-the-art models predominantly focus on high-resource languages such as English and Chinese, benefiting from abundant high-quality supervision and standardized benchmarks. In contrast, low-resource languages—particularly those prevalent in Southeast Asia (SEA)—continue to suffer from limited coverage, degraded supervision signals, and uneven performance across tasks. These limitations are further exacerbated in specialized domains such as e-commerce, where data is semi-structured, entity-centric, and distributionally distinct from general web text. As a result, general-purpose embedding models often fail to provide reliable semantic representations for multilingual e-commerce applications, leading to suboptimal retrieval, recommendation, and search performance in practice.

Developing high-performance embeddings for such scenarios faces three fundamental challenges. **First**, large-batch contrastive training with in-batch negatives, the standard methodology for embedding model training, often suffers from false negatives. Notably, when the training batch size is sufficiently large, semantically related documents from the same latent class (e.g. belonging to the same category or source document) are often inadvertently sampled for use as negatives, creating noisy gradients that penalize correct similarity identification and degrade semantic alignment, introducing artifacts that distorts representation geometry and slows convergence. **Second**, the scarcity and heterogeneity of training data in low-resource languages make it difficult to learn well-aligned semantic spaces. Available data is often noisy, weakly supervised, or misaligned with downstream retrieval objectives, rendering naive data scaling ineffective. **Third**, production deployment imposes strict constraints on inference latency and throughput, requiring embedding models to balance representational power with computational efficiency, often under aggressive quantization and serving optimizations.

We present Compass-Embedding v4, a multilingual e-commerce embedding model addressing these challenges through three synergistic innovations:

1. We introduce **Class-Aware Masking (CAM)**, a lightweight yet principled modification to the InfoNCE objective that mitigates false-negative bias in large-batch contrastive learning. By leveraging class-level metadata to selectively mask invalid in-batch negatives, CAM improves semantic discrimination without altering training complexity or scalability. To further stabilize domain adaptation, we employ spherical model merging to reconcile task-specialized representations with the original base model, effectively mitigating catastrophic forgetting.

2. We address data scarcity and imbalance through a diversified data construction pipeline that combines public multilingual corpora, context-grounded synthetic data generation, cross-lingual translation, and large-scale proprietary e-commerce data. This strategy emphasizes coverage and contrastive utility over raw volume, enabling robust semantic learning in low- and medium-resource languages while preserving generalization to high-resource languages.

3. We design Compass-Embedding v4 with production deployment in mind. Leveraging large-batch robustness, inference-aware training, and system-level optimizations—including vLLM-based serving, prefix caching, and FP8 quantization—we achieve high-throughput, low-latency inference while maintaining embedding quality. This enables Compass-Embedding v4 to operate as a cost-efficient alternative to proprietary embedding services in large-scale industrial settings.

We conduct extensive evaluations across multilingual public benchmarks and proprietary e-commerce tasks, covering retrieval, classification, clustering, and bitext mining. Results demonstrate that Compass-Embedding v4 achieves state-of-the-art performance on major Southeast Asian languages and delivers substantial gains on core e-commerce retrieval tasks, while maintaining competitive performance on high-resource language benchmarks. These findings highlight the effectiveness of our approach in bridging the performance gap between general-purpose embedding models and the practical requirements of multilingual e-commerce systems.

The remainder of this report is organized as follows. Section 2 describes the model architecture and design choices. Sections 3 and 4 detail the training methodology and data construction pipeline. Section 5 presents inference optimization strategies for production deployment. Section 6 reports experimental results across multilingual and domain-specific benchmarks. Finally, Section 7 concludes with a discussion of limitations and future directions.

## 2. Model Architecture

We adhere to the standard architectural paradigm for LLM-based embedding models, utilizing a **decoder-only Transformer block** where the conventional language modeling head is replaced by a pooling layer.

### 2.1. Causal Attention & Pooling

A critical distinction in our design is the use of **causal (unidirectional) attention**. Unlike traditional embedding models (e.g., BERT-based and most LLM encoders) that utilize *bidirectional attention* to access context from both left and right simultaneously, our model inherits the autoregressive nature of its LLM backbone. In this setup, each token can only attend to its predecessors.

Consequently, to capture the semantic representation of the entire sequence, we must employ **Last Token Pooling** (as illustrated in Figure 2). We append a special end-of-sequence token ([EOS]) to the input, and the embedding vector is derived directly from its final hidden state ($H_{\texttt{[EOS]}}$), which represents the aggregated context of the full sequence.

### 2.2. Instruction-Aware Inputs

Our model is designed to be instruction-aware, allowing users to explicitly define the intent of their embedding task. We format our instructions using the following template:

```
Instruction: {instruction}\nQuery:
```

**Instructions are applied** to **all embedding inputs except for documents in retrieval and re-ranking tasks**. For instance, a typical query is formatted as follows:

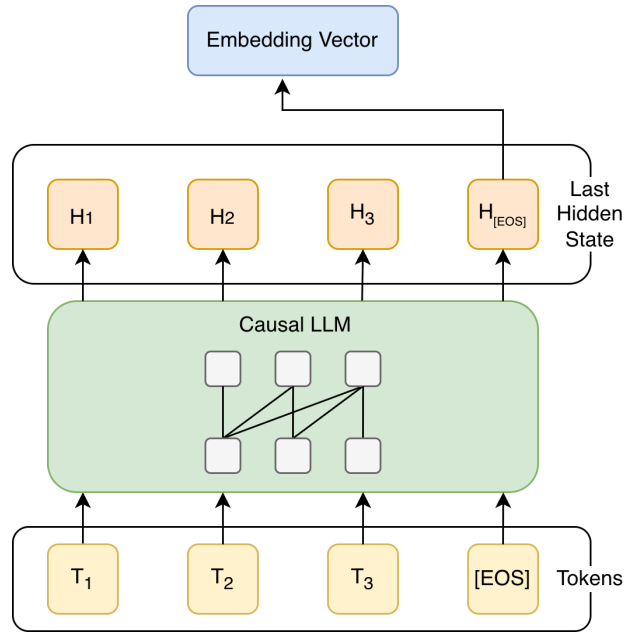

Figure 2 | Architecture of our model.

```
Instruction:  Given a web search query, retrieve relevant passages
that answer the query\nQuery:  Explain the definition of gravity
```

Conversely, our model is optimized to be **instruction-agnostic** for **documents used as retrieval or re-ranking candidates**. This allows such documents to be indexed in their unmodified state without limiting them to only be used for certain tasks. For instance, a retrieval candidate is embedded simply as:

```
Gravity is a force that attracts two bodies towards each other.  It
gives weight to physical objects and is responsible for the movement
of planets around the sun.
```

During training, we **apply instructions only to the queries**, leaving positive and negative documents to be embedded verbatim.

## 3. Model Training

We trained our embedding model on a comprehensive multilingual and e-commerce corpus utilizing a contrastive learning framework. Specifically, we adopt the InfoNCE loss objective (He et al., 2020), a standard in modern representation learning, which optimizes the vector space by minimizing the distance between query-positive pairs while maximizing the distance to negative candidates. To overcome memory constraints and scale the effective batch size to 1024, we employ GradCache (Gao et al., 2021). Furthermore, to enhance domain adaptation and parameter efficiency, we integrate specialized techniques including Class-Aware Masking (Section 3.1), and Model Merging (Section 3.2).

## 3.1. Training Objective with Class-Aware Masking

While scaling the batch size via GradCache improves the contrastive signal, it simultaneously increases the probability of encountering False Negatives. In a large batch, it is statistically probable that multiple distinct samples will originate from the same latent semantic class (e.g., sharing the same classification label or generated from the same expert persona). In the standard InfoNCE formulation, these semantically valid matches are treated as negatives, introducing noisy gradients that punish the model for correctly identifying similarity. We argue that false negatives in large-batch contrastive learning are not merely noise, but a systematic bias induced by heterogeneous supervision sources. Class-aware masking provides a lightweight yet principled correction without altering the core InfoNCE objective.

To mitigate this, we introduce a Class-Aware Masking mechanism. We assign a discrete class identifier to every training instance, derived from its metadata (such as category labels or synthetic generation prompts). During the loss computation, we dynamically mask the interaction between any query and in-batch negative that share the same class identifier.

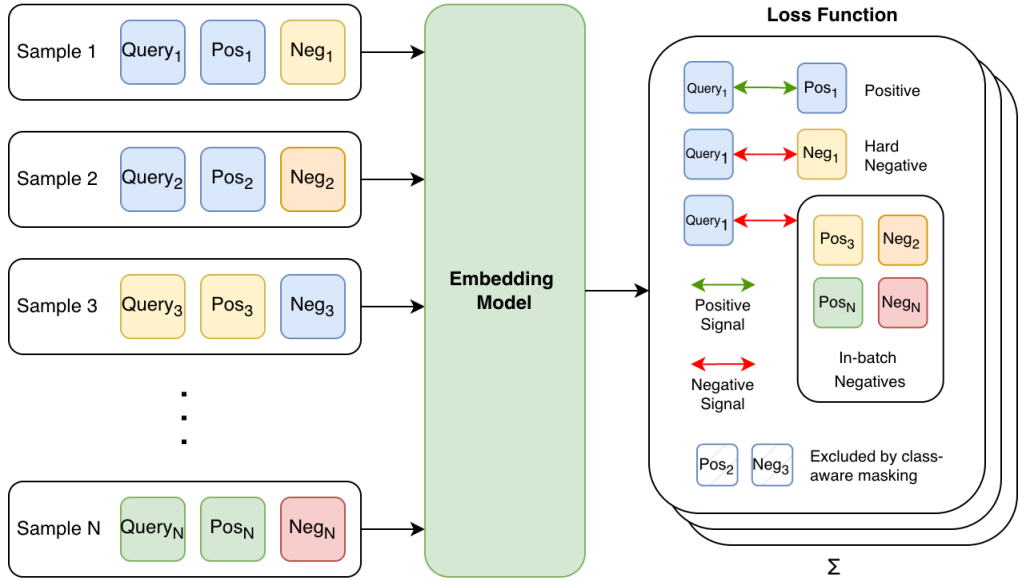

Figure 3 | Illustration of our class-aware masking strategy. Content of the right-hand side box depicts computation of $\mathcal{L}_1$ (Equation 2), with the color of each square denoting the class of each document. Note that the positive document from the second sample ($d_2^+$), and negative document from the third sample ($d_3^-$) are not used as in-batch negatives for query $q_1$ because their classes ($c_2^+, c_3^-$) are the same as that of $q_1$ ($c_1^+$).

Formally, for a batch of $N$ query–document pairs, the InfoNCE loss is computed with binary masks applied to the denominator, preventing invalid in-batch comparisons. This modification preserves the computational efficiency and scalability of standard contrastive learning, while substantially reducing gradient noise caused by false negatives. The overall loss for a batch is defined as:

$$\mathcal{L} = \frac{1}{N} \sum_{i=1}^{N} \mathcal{L}_i \tag{1}$$

where $\mathcal{L}_i$ is the loss for a single query $q_i$:

$$\mathcal{L}_i = -\log \frac{\exp\left(\sigma(f_\theta(q_i), f_\theta(d_i^+))\right)}{D_i} \tag{2}$$

The denominator $D_i$ aggregates similarities between query $q_i$ and all documents in the batch:

$$D_i = \sum_{j=1}^{N} m_{ij}^+ \cdot \exp\left(\sigma(f_\theta(q_i), f_\theta(d_j^+))\right) + \sum_{j=1}^{N} m_{ij}^- \cdot \exp\left(\sigma(f_\theta(q_i), f_\theta(d_j^-))\right) \tag{3}$$

The masking rules $m_{ij}^+$ and $m_{ij}^-$ are defined as:

$$m_{ij}^+ = \begin{cases} 0, & \text{if } c_i^+ = c_j^+ \text{ and } i \neq j \\ 1, & \text{otherwise} \end{cases} \tag{4}$$

$$m_{ij}^- = \begin{cases} 0, & \text{if } c_i^+ = c_j^- \\ 1, & \text{otherwise} \end{cases} \tag{5}$$

Where the symbols are defined as follows:

| Symbol | Definition |
|--------|------------|
| $N$ | Number of queries in a batch |
| $q_i$ | Query of sample $i$ |
| $d_i^\pm$ | Positive (+) and negative (−) documents of sample $i$ |
| $c_i^\pm$ | Class identifiers for documents $d_i^\pm$ |
| $\sigma(\cdot)$ | Cosine similarity function |
| $f_\theta(\cdot)$ | Embedding model parameterized by $\theta$ |

## 3.2. Model Merging

Post-training evaluation revealed a performance trade-off: while our intermediate model achieved significant gains in Southeast Asian languages and E-commerce contexts, it suffered regression in general performance indicators such as English benchmark performance. To mitigate this forgetting, we adopted a weight merging strategy to combine checkpoints from earlier and later in our training process, drawing on methodologies from Zhang et al. (2025) and Li et al. (2024).

We specifically employed Spherical Linear Interpolation (SLERP) (Shoemake, 1985) for this fusion. From a representation geometry perspective, naive parameter averaging distorts embedding space isotropy. By interpolating along a hyperspherical geodesic, SLERP preserves angular relationships critical for similarity-based retrieval. The resulting fusion successfully recovered performance in the general domains while retaining the specialized gains, yielding a robust and balanced final model.

## 4. Training Data

High-quality and diverse training data is the cornerstone of an effective embedding model. To achieve the required level of fidelity and coverage, we constructed a comprehensive training corpus of 3.57 million samples. Rather than maximizing raw data volume, our data construction prioritizes coverage, alignment, and contrastive utility, especially for low-resource languages where naïve scaling is ineffective. This dataset is synthesized from three distinct sources: public corpora, synthetic data generated by Large Language Models (section 4.2.2), and proprietary business data from our e-commerce operations (section 4.3). To maximize the utility of these sources, we employ extensive data processing techniques designed to improve the efficiency of the training signal (section 4.1) and bridge coverage gaps prevalent in multilingual domains (section 4.2).

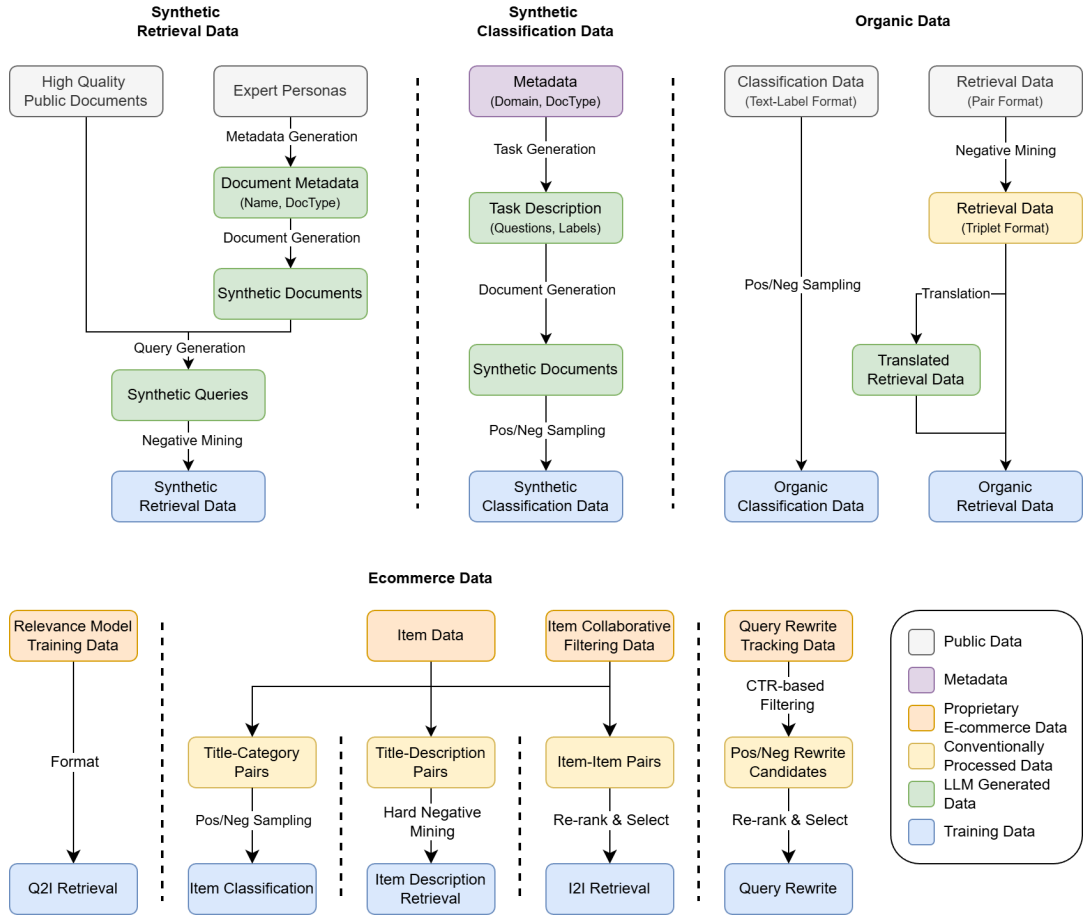

Figure 4 | Data pipelines of Compass-Embedding v4.

## 4.1. Data Augmentation Strategies

Contrastive triplets (Query, Positive, Negative) rarely occur naturally in raw data; consequently, the majority of our training corpus is produced from standard query-positive pairs. Nevertheless, training signal provided by a well-curated negative component is essential in training a high performance embedding model (Xiong et al., 2020). We utilize a task type dependent negative selection method inspired by those used by recent state-of-the-art embedding models. Classification tasks utilize class-based sampling to convert query-label pairs into triplets (Lee et al., 2025), whereas retrieval tasks rely on percentage margin based hard negative mining (de Souza P. Moreira et al., 2025) to match query-positive pairs with suitable negatives.

## 4.2. Data Expansion Strategies

To address the inherent data scarcity of low-resource languages, we augment our training corpus with translated and synthesized data. These approaches contribute to the diversification of our dataset, closing critical coverage gaps and improving the model's capabilities among these low resource languages.

### 4.2.1. Data Translation

Low-resource language training corpus frequently suffer from lack of comprehensiveness, leading to distinct capability gaps in trained models. For instance, preliminary evaluations revealed that our model initially demonstrated suboptimal performance in Wikipedia passage retrieval for Southeast Asian (SEA) languages—a task that is typically robust in English-centric models.

To mitigate this deficit, we employed a cross-lingual transfer strategy, hypothesizing that high-resource capabilities could be projected into the target languages via translation. We utilized **GPT-5** to translate the **SQuAD** (Rajpurkar et al., 2016) dataset, a standard dataset on English Wikipedia retrieval, into the target SEA languages. The integration of this translated corpus into our training pipeline proved effective: subsequent evaluations showed a marked improvement in multilingual retrieval capabilities, yielding performance improvements of 1% or greater across the majority of downstream tasks.

### 4.2.2. Data Synthesis

Synthetic data generation is an established paradigm for overcoming data scarcity and enhancing domain diversity. However, naive generation methods often yield unnatural artifacts that fail to translate into effective training signals. To circumvent this, we implement three distinct generation pipelines tailored to each of the 3 common task types of retrieval, binary classification, and multiclass classification.

A critical differentiator in our approach is contextual grounding. Rather than relying on unconstrained or weakly constrained generation method often found in earlier works such as Wang et al. (2024), we condition our generative model on "grounding data" consisting of curated expert personas and high-quality seed passages. This constraint ensures that the resulting training samples are complex, diverse, and realistic.

- **Expert Text Generation**: Inspired by Qwen3 (Zhang et al., 2025), our pipeline adopts the expert personas from Ge et al. (2025) to guide the synthesis of diverse documents. Our method introduces a two-stage refinement: first, the LLM is prompted with a persona to generate a list of potential document names and types (e.g., technical report, blog post, magazine article). Subsequently, the LLM is prompted again to generate the full content of each document. For lengthy documents exceeding standard generation limits, our pipeline iteratively generated content chapter by chapter.
- **Retrieval Dataset Construction**: The generated expert passages are combined with high-quality text from public sources like Nemotron-CC (Su et al., 2025) and Cosmopedia (Ben Allal et al., 2024; Dou et al., 2025) to form a comprehensive corpus. This corpus is then fed into an LLM to generate 5 queries per document, leveraging the detailed content to produce diverse queries covering various aspects of a passage. The final step in this sub-pipeline is hard negative mining. To mitigate the false negative problem, each text is assigned a class identifier: Expert Texts receive an identifier unique to their persona, while each unique document in the General Texts receive a unique identifier. A condition is applied to ensure that no samples have its positive and negative documents share the same identifier. We also integrated a smaller, legacy dataset generated following Muennighoff et al. (2025).
- **Classification Dataset Construction**: The framework introduces a novel, multi-step generation process for classification tasks, conditioned on language, domain (e.g., Finance, Sports, Science), and document type (e.g., Technical Documentation, Advertisement). For each combination, an LLM generates a list of plausible classification tasks.
  - For **binary classification**, the pipeline follows a three-step process: (1) generate the task definition and label descriptions, (2) generate documents for each label, and (3) format each document into a triplet using itself as the anchor, its corresponding label text as the positive, and the opposite label text as the negative.

– For **multiclass classification**, a four-step process is used: (1) generate the task, (2) generate the set of class options, (3) generate documents for each class, and (4) form triplets using a document as the anchor, another document from the same class as the positive, and a document from a different class as the negative.

### 4.3. E-commerce Data Construction

| ID | Task Name | Entities | Goal | Source |
|----|-----------|----------|------|--------|
| T1 | Query to Item Retrieval | Query ↔ Item | Retrieve items relevant to the user's query | Training data for Query-Item Relevance (QIR) model |
| T2 | Item to Item Retrieval | Item ↔ Item | Retrieve items related to the given item (e.g. YMAL page) | Retrieve top 128 candidates using collaborative filtering, and re-rank using a semantic re-ranking model |
| T3 | Category Prediction | Item ↔ Category | Classify items into L1 and L2 categories based on its title | Item table; formatted into triplet using the method in section ?? |
| T4 | Query Rewrite | Query ↔ Query | Rewrite a search query into a different but synonymous one to boost number of recalled items | Historical traffic data of query-rewrite pairs; ranked by combination of CTR and semantic re-ranking model results |
| T5 | Item Description Retrieval | Item Title ↔ Item Description | Retrieve the full item description given its title | Item data table; hard negative is chosen by hard negative mining |

Table 1 | Summary of E-Commerce Tasks and Dataset Construction

General text embedding models often struggle to generalize to e-commerce environments due to fundamental differences in data format and content. Unlike general text, e-commerce data is often semi-structured, entity-dense, and highly specific. Therefore, generic datasets are insufficient; the model requires targeted exposure to e-commerce entities and their relationships to learn effectively.

To facilitate this, we introduce a dedicated e-commerce dataset consisting of five specific tasks. These tasks were selected to cover a wide variety of entities (e.g., Queries, Items, Categories) and relationships (e.g., Query-Item, Query-Query, Query-Category). All samples were constructed using in-house data collected from our large-scale e-commerce operations. Description and construction method of each tasks can be found in table 1. E-commerce embedding fundamentally differs from generic text embedding in that it encodes typed entities and typed relations. Our task suite explicitly covers Query–Item, Item–Item, and Query–Query relations, enabling the model to learn a structured semantic space rather than flat textual similarity.

### 4.4. Training Dataset Composition

**Task Type Distribution**: **Retrieval** constitutes 68% of the training data, reflecting the importance of RAG in our optimization goals. The balance is split between **Clustering** (13%), **Classification** (11%), and **NLI** (8%). We restricted Clustering and NLI solely to English samples. Multilingual clustering data was omitted due to scarcity and substitutability by classification data, while multilingual NLI

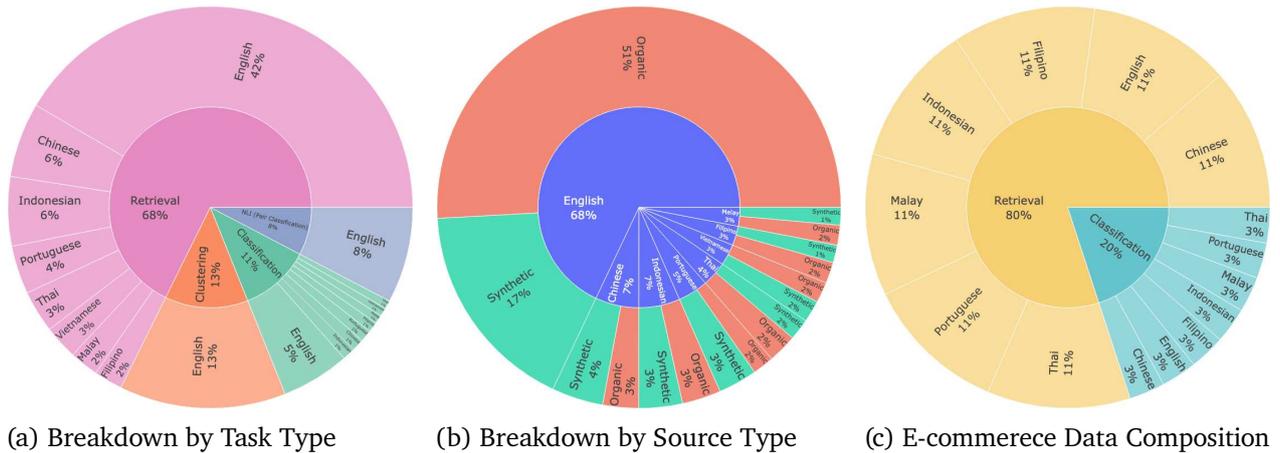

(a) Breakdown by Task Type     (b) Breakdown by Source Type     (c) E-commerece Data Composition

Figure 5 | Composition of our 3.57M samples training dataset. Diagrams (a) and (b) show the composition of our general domain dataset, while (c) shows the composition of our e-commerce dataset

was excluded after experiments showed its distinct training requirements negatively impacted global performance.

**Language Distribution**: Approximately **25%** of our general-domain corpus comprises samples from six low-to-medium resource languages. The specific allocation for each language is optimized based on a dual criterion: intrinsic data availability and projected downstream query volume. While synthetic data accounts for roughly **30%** of the global dataset, this proportion increases significantly to approximately **50%** within the lower resource group. This stratification underscores the role of synthetic data in mitigating data scarcity among these lower resource languages.

**E-commerce Data Distribution** We downsampled our e-commerce dataset to 250,880 samples, an amount chosen to provide sufficient data volume for robust domain adaptation without degrading general-purpose capabilities. This corpus is distributed evenly across five tasks, with each task assigned a quota of 50,176 to be equally subdivided between each languages. We treat the two (L1 and L2) item category classification tasks as a single unified task for the purpose of data volume allocation.

## 5. Inference Optimization

### 5.1. Cost Comparison

**Compass-Embedding v4 achieves an over 8× increase in inference throughput and an 88% reduction in cost relative to Compass-Embedding v3. Under identical service-level objective (SLO) constraints, the operational cost of Compass-Embedding v4 amounts to merely 15% of OpenAI's text-embedding-3-small.** The evaluation was conducted on a randomly sampled set of 10,000 real-world queries collected from our online embedding service. Both Compass-Embedding v3 and v4 were evaluated for throughput, latency, and cost under identical experimental settings on a single H100 GPU. To ensure a fair comparison with the OpenAI embedding API, we adopt the average observed online latency of OpenAI as the service SLO, and measure the achievable throughput and cost of Compass-Embedding v4 under this constraint. The results are presented in Figure 6.

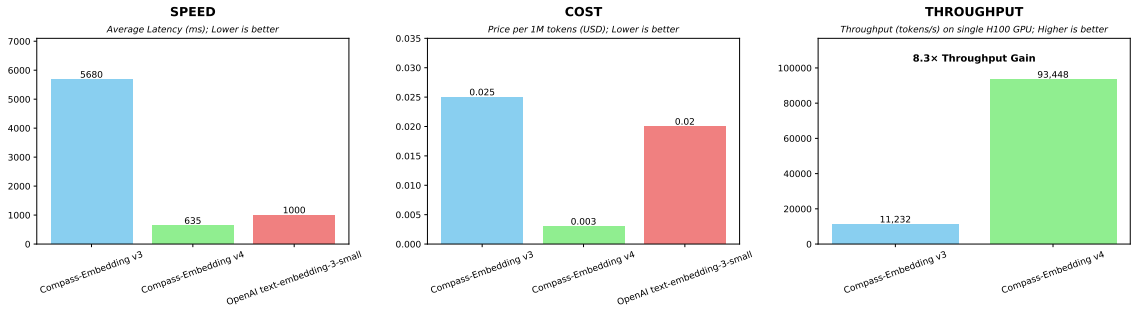

Figure 6 | Speed, Cost, and Throughput Evaluation of Compass-Embedding in Comparison with the OpenAI Embedding API. The throughput of the OpenAI embedding API is omitted, as it cannot be directly evaluated.

## 5.2. Acceleration Strategies

### 5.2.1. Bottlenecks

The embedding computation in Compass-Embedding v4 consists solely of prefill operations, with no autoregressive decoding. Unlike decoding-intensive tasks, embedding workloads are fully parallelizable across the input sequence. Profiling GPU utilization under varying request loads revealed that even at minimal traffic, utilization consistently exceeded **95%**, indicating that the workload is primarily **compute-bound**. Motivated by this, we investigated two complementary optimization strategies: Prefix Caching and FP8 Quantization.

### 5.2.2. Prefix Caching

Prefix caching stores the intermediate representations of frequently occurring input prefixes, avoiding redundant computation for subsequent requests that contain the same prefix. This technique is particularly effective for embedding tasks for two main reasons:

- **High input redundancy:** Real-world embedding workloads often contain repeated phrases, standard headers, or common prompt patterns. Caching these shared prefixes prevents recomputation of the same subsequences.
- **Compute-bound workload:** As indicated by our bottleneck analysis, embedding is dominated by GPU computation. Eliminating redundant calculations reduces the effective computational load, resulting in lower latency and higher throughput.

As shown in Table 2, evaluations conducted on real-world online traffic demonstrate that, with all other configurations held constant, prefix caching achieves a consistently high cache hit rate in the production embedding service. Specifically, the prefix cache hit rate reaches up to 49%, resulting in a substantial throughput improvement of 37.8%, along with noticeable reductions in end-to-end latency. These results indicate that prefix caching is highly effective in realistic deployment scenarios, where repeated or partially overlapping input prefixes are common.

### 5.2.3. FP8 Quantization

We employed **FP8 quantization** (W8A8) to further accelerate embedding inference. Two strategies were evaluated: **Fine-grained Per-Block FP8 Quantization:** Weights are quantized in $128 \times 128$ blocks with separate scales, and activations are dynamically quantized in $1 \times 128$ blocks to handle outliers. **Per-Tensor FP8 Quantization:** Weights are quantized with a single per-tensor scale, while activations are dynamically scaled per tensor during each forward pass. As shown in Table 2, **Per-Tensor FP8 quantization** outperforms the Per-Block approach in both inference throughput and average latency. This advantage stems from its coarser quantization granularity, which incurs lower computational overhead.

| Method | Cache Hit Rate (%) | Throughput (Tok/s) | Avg Latency (ms) | Throughput ↑ (vs BF16) | Latency ↓ (vs BF16) |
|---|---|---|---|---|---|
| BF16 Baseline | 0.00% | 65,099 | 1,364.77 | – | – |
| BF16 + Prefix Caching | 49.40% | 89,688 | 883.06 | 37.8% | 35.3% |
| Per-Block FP8 + Prefix Caching | / | 90,683 | 791.66 | 39.3% | 41.9% |
| Per-Tensor FP8 + Prefix Caching | / | 93,448 | 635.12 | 43.5% | 53.4% |

Table 2 | Cumulative throughput and latency improvements relative to the BF16 baseline. Per-block and per-tensor FP8 are evaluated as parallel configurations under prefix caching.

To further examine the impact on task-level performance, we evaluated multiple downstream benchmarks (Figure 7). Both FP8 quantization strategies exhibited minimal performance degradation: compared to BF16, Per-Tensor FP8 incurred an average drop of **0.13%**, while Per-Block FP8 resulted in **0.09%**. Given the negligible difference in accuracy and the higher efficiency of per-tensor quantization, we adopted the per-tensor approach for production deployment.

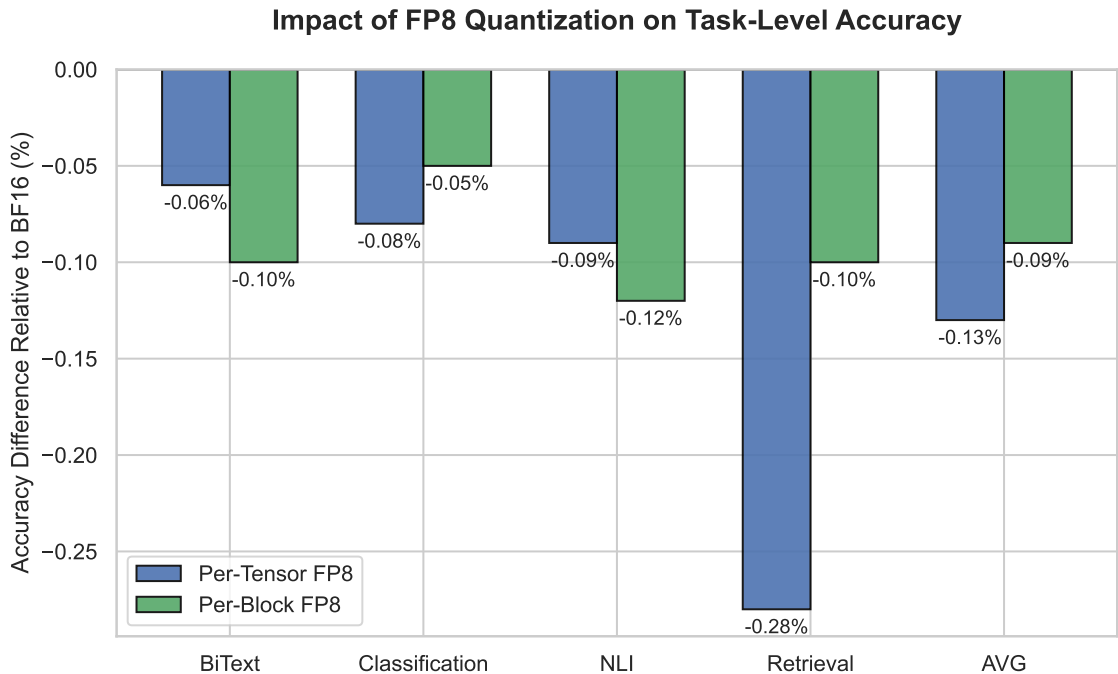

Figure 7 | Accuracy Impact of FP8 Quantization Compared to BF16 Across Downstream Tasks

## 6. Results

We trained our model, **Compass-Embedding v4**, using LoRA parameter efficient fine-tuning (Hu et al., 2022). To rigorously assess the capabilities of our model, we conducted a comprehensive evaluation across three distinct axes: low-resource language alignment (Southeast Asian languages and Portuguese), domain-specific capabilities (E-commerce), and generalist performance (English).

In the following sections, we benchmark our model against a robust set of baselines, including our previous generation model (Compass-Embedding v3), and state-of-the-art proprietary and open-source models such as Qwen3-Embedding-4B, OpenAI's Text-Embedding-3 series and BGE-Multilingual-Gemma2. This multi-faceted evaluation strategy ensures that our model not only excels in its target niche but also maintains competitive robustness in general applications. Instruction prompts for our SEA languages and E-commerce benchmarks can be found in tables 13 and 14, while English tasks use the same instructions as those used by *Qwen3-Embedding* series.

| Model | Size | Weighted | Task Type | | | |
|---|---|---|---|---|---|---|
| | (B) | Average | Retrv. (21) | Classf. (18) | Bitext (12) | NLI (4) |
| *Our Models* | | | | | | |
| **Compass-Embedding v4** | 4 | **84.26** | <u>85.12</u> | **79.06** | 96.19 | <u>67.39</u> |
| Compass-Embedding v3 | 7 | 78.31 | 80.16 | 70.75 | 91.43 | 63.27 |
| *Proprietary Models* | | | | | | |
| Text-Embedding-3-Large | - | 76.24 | 80.19 | 64.86 | 93.06 | 56.30 |
| Text-Embedding-3-Small | - | 66.53 | 72.12 | 58.01 | 73.42 | 54.90 |
| *Open Source Models* | | | | | | |
| BGE-Multilingual-Gemma2 | 9 | <u>83.16</u> | **85.90** | 74.59 | <u>96.49</u> | **67.40** |
| Qwen3-Embedding-4B | 4 | 82.89 | 84.17 | <u>76.41</u> | 95.61 | 67.21 |
| F2LLM-4B | 4 | 80.03 | 80.20 | 75.49 | 93.21 | 60.04 |
| Multilingual-E5-Large-Instruct | 0.56 | 77.95 | 81.67 | 65.61 | **96.52** | 58.27 |
| KaLM-Multilingual-Mini-Instruct-v2.5 | 0.5 | 73.91 | 77.81 | 70.42 | 77.93 | 57.10 |

Table 3 | SEA Languages and Portuguese Evaluation Results

## 6.1. Southeast Asian Languages and Portuguese Benchmark

To assess the model's capabilities in Southeast Asian languages and Portuguese, we employed a comprehensive multilingual benchmark comprising 19 distinct tasks. This evaluation suite spans four primary categories: Retrieval, Classification, Bitext Mining, and Natural Language Inference (NLI).

The benchmark was constructed by curating tasks from the Massive Text Embedding Benchmark (MTEB) ecosystem. Our selection protocol prioritized tasks that are already established in the standard Multilingual MTEB leaderboard to ensure quality. Furthermore, to guarantee a balanced composite metric, we favored datasets with broad cross-lingual coverage (e.g., BelebeleRetrieval) and strictly controlled the distribution to ensure equal task representation across all target languages.

Evaluation results in this category demonstrate that **Compass-Embedding v4** (84.26) achieves consistent state-of-the-art performance. It surpasses both foundational **Qwen3-Embedding-4B** (82.89) and other strong open-source baselines, exhibiting particular dominance in classification tasks and overall weighted average scores. Furthermore, the model establishes a significant lead over proprietary industry standards, outperforming OpenAI's **Text-Embedding-3-Large** (76.24) and **Text-Embedding-3-Small** (66.53) by margins of **10.5%** and **26.6%**, respectively.

| Model | Size | Avg. | Task Type | | | | | |
|---|---|---|---|---|---|---|---|---|
| | (B) | | ItemL1 | ItemL2 | I2I | ItemDesc. | Q2I | Q2Q |
| *Our Models* | | | | | | | | |
| **Compass-Embedding v4** | 4 | **68.39** | **71.41** | <u>55.01</u> | **76.74** | **83.24** | **33.24** | **90.71** |
| Compass-Embedding v3 | 7 | 58.87 | 62.75 | 49.12 | 71.74 | 74.28 | 19.40 | 75.93 |
| *Proprietary Models* | | | | | | | | |
| Text-Embedding-3-Large | - | 61.39 | 57.24 | 46.54 | <u>74.85</u> | 77.62 | 25.03 | 87.09 |
| Text-Embedding-3-Small | - | 55.75 | 52.04 | 40.90 | 68.43 | 69.87 | 19.79 | 83.51 |
| *Open Source Models* | | | | | | | | |
| BGE-Multilingual-Gemma2 | 9 | <u>63.56</u> | 61.74 | **55.96** | 73.68 | <u>79.24</u> | 21.88 | <u>88.85</u> |
| Qwen3-Embedding-4B | 4 | 62.35 | <u>63.77</u> | 50.17 | 70.56 | 76.59 | <u>27.94</u> | 85.09 |
| F2LLM-4B | 4 | 52.83 | 54.15 | 43.18 | 62.53 | 67.50 | 15.73 | 73.90 |
| Multilingual-E5-Large-Instruct | 0.56 | 54.98 | 50.52 | 38.72 | 73.82 | 73.73 | 10.70 | 82.41 |
| KaLM-Multi-Mini-Inst-v2.5 | 0.5 | 56.15 | 54.05 | 37.89 | 70.44 | 69.41 | 23.03 | 82.10 |

Table 4 | Ecommerce Evaluation Results

## 6.2. E-commerce Results

We further validated the model's performance on the held-out test split of our proprietary e-commerce dataset, which encompasses six distinct tasks ranging from Item-to-Item (I2I) recommendation to Query-to-Item (Q2I) retrieval. As detailed in Table 4, Compass-Embedding v4 achieved an average score of 68.39, demonstrating that the model has successfully internalized the e-commerce domain concepts.

Quantitatively, this represents a 16.2% generation-over-generation improvement relative to our previous Compass-Embedding v3 model (58.87). Furthermore, our model establishes a new baseline for this domain, outperforming Qwen3-Embedding-4B (62.35) by 9.7%, and surpassing the commonly used Text-Embedding-3 series of model from OpenAI by more than 10%.

## 6.3. English Results

While our optimization efforts were primarily directed toward Southeast Asian languages and e-commerce scenarios, maintaining robustness in high-resource languages is essential for a versatile embedding model. To evaluate these "generalist" capabilities, we benchmarked Compass-Embedding v4 on the English portion of the Massive Text Embedding Benchmark (MTEB) collection.

For this category, Compass-Embedding v4 achieved an average score of **74.30** (Table 5). This score surpasses all comparable open-source models with the exception of Qwen3-Embedding-4B (74.55), where the performance difference is statistically negligible (< 0.3%). Furthermore, our model significantly outperforms proprietary industry standards, surpassing Text-Embedding-3-Large (66.49) by nearly 8 points and achieving a 6.7-point improvement over our previous generation model. This outcome demonstrates the success of our **SLERP parameter fusion** technique, which allowed us to integrate significant volumes of domain-specific data without triggering catastrophic forgetting or compromising general capabilities.

| Model | Size | Avg. | Task Type | | | | | | |
|---|---|---|---|---|---|---|---|---|---|
| | (B) | | Cls. | Clust. | NLI | Rrk. | Ret. | STS | Sum. |
| *Our Models* | | | | | | | | | |
| **Compass-Embedding v4** | 4 | 74.30 | 90.53 | 58.56 | 87.04 | 50.15 | 66.68 | 87.64 | 36.77 |
| Compass-Embedding v3 | 7 | 67.62 | 80.66 | 54.21 | 85.76 | 48.56 | 55.18 | 83.41 | 36.65 |
| *Proprietary Models* | | | | | | | | | |
| Text-Embedding-3-Large | - | 67.33 | 79.13 | 48.90 | 85.82 | 48.68 | 57.98 | 81.44 | 34.31 |
| Text-Embedding-3-Small | - | 64.62 | 77.56 | 47.46 | 85.06 | 47.29 | 53.48 | 81.35 | 32.49 |
| *Open Source Models* | | | | | | | | | |
| BGE-Multilingual-Gemma2 | 9 | 73.65 | 88.72 | 58.9 | 85.97 | 48.19 | 61.82 | 84.22 | **37.47** |
| Qwen3-Embedding-4B | 4 | **74.55** | 89.83 | 57.50 | **87.37** | **50.32** | **68.15** | **88.72** | 35.37 |
| F2LLM-4B | 4 | 73.67 | **91.68** | **68.54** | 83.75 | 50.05 | 59.63 | 84.2 | 33.19 |
| Multilingual-E5-Large-Instruct | 0.56 | 65.53 | 75.54 | 49.89 | 86.24 | 48.74 | 53.47 | 84.72 | 29.89 |
| KaLM-Multi-Mini-Inst-v2.5 | 0.5 | 71.29 | 90.50 | 58.12 | 86.63 | 47.42 | 58.45 | 84.82 | 31.21 |

Table 5 | MTEB(English, v2) Evaluation Results

## 7. Conclusion

In this work, we introduced Compass-Embedding v4, a multilingual embedding model designed for large-scale e-commerce applications with a particular focus on Southeast Asian languages. By integrating class-aware contrastive learning, a diversified multilingual and domain-specific data pipeline, and deployment-aware optimizations, Compass-Embedding v4 achieves strong and balanced performance across low-resource languages, e-commerce retrieval tasks, and general-purpose benchmarks. Extensive experiments demonstrate that the proposed approach effectively mitigates false-negative bias in contrastive learning, improves cross-lingual and domain alignment, and remains highly efficient under production serving constraints.

Looking ahead, Compass-Embedding can be further extended along several promising directions. These include more fine-grained product representations, multimodal embeddings that integrate textual and visual signals, deeper and more robust multilingual alignment, and embedding representations enhanced with lightweight reasoning capabilities. We believe that continued research in these directions will further strengthen the role of embedding models as foundational components for next-generation multilingual retrieval, recommendation and RAG systems.

Ultimately, we hope that Compass-Embedding v4 proves to be a practical utility for real-world applications in multilingual e-commerce and beyond; and to serve as a significant step forward in closing the performance gap for under-represented languages and applications.

## A. Prompts For Synthetic Data Generation

### A.1. Expert Text Generation

See tables 6 and 7.

### A.2. Synthetic Query Generation

See table 8

### A.3. Binary Classification Data Generation

See tables 9 and 10.

### A.4. Multiclass Classification Data Generation

See tables 11 and 12.

## B. Benchmark Datasets

See tables 13 and 14

| **Step 1 (Task Generation)** |
|---|

You are an expert with the following persona:

{persona}

List 5 different types of short text (less than 250 words) you might be able to write. For each type, list the following:
1. The title of the content, which should be in {language}
2. The medium in which to write the content, which must be textual in nature

Return the results in JSON format.

| **Step 2 (Document Generation)** |
|---|

You are an expert with the following persona:

{persona}

Write a {medium} on "{title}". The {medium} should be in {language}. Be specific, do not use placeholders for names, emails, and other named entities. Do not output anything else.

Table 6 | Expert Text (Short Variant)

---

**Step 1 (Task Generation)**

---

You are an expert with the following persona:

{persona}

List 3 different types of long text (more than 250 words) you might be able to write. For each type, list the following:

1. The title of the content.
2. The medium in which to write the content, which must be textual in nature
3. List of sections

The title and list of sections should be in
{language}.

Return the results in json format.

---

**Step 2 (Document Generation)**

---

You are an expert with the following persona:

{persona}

Write the "{section}" section of a {medium} on "{title}". Do not output anything else. Since this is a part of a longer document, you should end abruptly without providing a conclusion.

---

Table 7 | Expert Text (Long Variant)

---

**Prompt**

You are an annotator who is building a dataset to train a text embedding model. You will be given a document, come up with 5 scenarios of various types where a user's query leads to the document. These scenarios may include themes such as finding answers to a question, searching for further information or content of a certain type on a topic, or to find evidence to support or refute a claim.

For each scenario, generate the following:

- The background of who the user is and what they are trying to do
- The generic instruction string that sets the retrieval objective. This should be drawn from a list of fixed templates
- The actual query written by the user. Ensure variation in length across scenarios.

The following are some sample instructions and queries:

...

The instruction must be as generic as possible. It must not contain topics discussed in the document. Pay attention to the instruction text when writing the query. The query should also not repeat long sections of the text verbatim. Each query you write should also vary in length. Return your result in json format.

The instruction must be in English, while the query must be in {language}.

Document:
{document}

---

Table 8 | Synthetic Query Generation

---

**Step 1 (Task Generation)**

You are an annotator working on synthetic data to build a classification model. Brainstorm 10 binary classification tasks of various types (e.g. content classification, reasoning, sentiment classification) where the input is a {doctype} in {domain}. Return the results as a json object, including the task instruction and a short but unambiguous label name for the positive and negative options. The instruction should be in English while label names should be in {language}. If the label names already exists in the instruction, you should use it. Be creative.

---

Table 9 | Binary Classification Task Generation

---

**Step 2 (Text Generation - Short Variant)**

---

You are an annotator working on synthetic data to build a classification model. Given the following binary classification task to classify a {doctype} in the in domain domain: "{task}" Write {count} {doctype} each for the positive label "{positive}", and negative label "{negative}". The {doctype} should be written in {language}. Be specific, and include names where appropriate. Return the result as json.

---

**Step 2 (Text Generation - Long Variant)**

---

You are an annotator working on synthetic data to build a classification model. Given the following binary classification task to classify a {doctype} in the in domain domain: "{task}" Write {count} {doctype} each for the positive label "{positive}", and negative label "negative". The {doctype} should be written in {language}. Be detailed and specific, and include names where appropriate. Return the result as json.

---

Table 10 | Binary Classification Text Generation

---

**Step 1 (Task Generation)**

---

You are an annotator working on synthetic data to build a classification model. Brainstorm 10 classification task prefix instructions to classify a {doctype} in {domain} domain. Focus on task with large number of classes. Return the results as json. Do not indicate any classes or options. Be specific but creative. Each task should only classify the text based on a single dimension. Do not output task numbers.

---

**Step 2 (Options Generation)**

---

Given the following classification task to classify a {doctype} on {domain}: "{task}". Determine if it is possible to generate valid options within the {domain} domain. If it is not, return an empty list. Otherwise, generate as many distinct options as reasonable. Return your result in json format.

---

Table 11 | Multiclass Classification Task & Options Generation

**Step 3 (Text Generation - Short Variant)**

You are an annotator working on synthetic data to build a classification model. Given the following classification task to classify {doctype} on {domain}: "task" Write count short {doctype} to be used as training samples. The class of this {doctype} should be "{option}". The {doctype} should be written in {language}. Avoid repeating the class name in the text as much as possible. Be specific, and include names where appropriate. The class of these {doctype} should be "{option}". Return your results as json.

**Step 3 (Text Generation - Long Variant)**

You are an annotator working on synthetic data to build a classification model. Given the following classification task to classify a {doctype} on {domain}: "task" Write a {} to be used as a training sample. The class of this text should be "{option}". Be detailed and specific, and include names where appropriate. Avoid repeating the class name in the text as much as possible. The {doctype} should be written in {language}. Output the {doctype} only and nothing else.

Table 12 | Multiclass Classification Text Generation

| Dataset Name | Supported Languages | Prompt |
|---|---|---|
| **Retrieval** | | |
| BelebeleRetrieval | ID, MS, PT, TH, TL, VI | Retrieval the relevant passage for the given query |
| MIRACLRetrievalHardNegatives | ID, TH | Retrieval relevant passage for the given query |
| MLQARetrieval | VI | Retrieval the relevant passage for the given query |
| MrTidyRetrieval | ID, TH | Retrieval relevant passage for the given query |
| WebFAQRetrieval | ID, MS, PT, TH, TL, VI | Given a question, retrieve passages that answer the question |
| WikipediaRetrievalMultilingual | PT | Retrieval relevant passage for the given query |
| XPQARetrieval | PT | Given a web search query, retrieve relevant passages that answer the query |
| XQuADRetrieval | TH, VI | Given a query, retrieve the answer to that query |
| **Classification** | | |
| FilipinoShopeeReviewsClassification | TL | Given a shop review, classify its rating on a scale from 1 to 5 |
| IndonesianIdClickbaitClassification | ID | Given an Indonesian news headlines, classify its into clickbait or non-clickbait |
| MongabayConservationClassification | ID | Represent the passage for finding another passage with the same sentiment (positive / neutral / negative) |
| MassiveIntentClassification | ID, MS, PT, TH, TL, VI | Given a user utterance as query, find the user intents |
| MassiveScenarioClassification | ID, MS, PT, TH, TL, VI | Given a user utterance as query, find the user scenarios |
| MultilingualSentimentClassification | ID, TH, VI | Given a text, categorized by sentiment into positive or negative |
| **Bitext Mining** | | |
| BibleNLPBitextMining | ID, MS, PT, TH, TL, VI | Retrieve parallel sentences |
| Tatoeba | ID, MS, PT, TH, TL, VI | Retrieve parallel sentences |
| **Natural Language Inference (NLI)** | | |
| IndoNLI | ID | Retrieve semantically similar text |
| SICK-BR-PC | PT | Represent the sentence to find another sentence with the same meaning |
| XNLI | TH, VI | Retrieve semantically similar text |

Table 13 | SEA Languages and Portuguese Evaluation Datasets

| Dataset Name | Supported Languages | Prompt |
|---|---|---|
| **Retrieval** | | |
| EcomI2IRetrieval | ID, MS, PT, TH, TL, VI, ZH | Given an ecommerce item, retrieve a related item |
| EcomItemDescriptionRetrieval | ID, MS, PT, TH, TL, VI, ZH | Given an ecommerce item title, retrieve its description |
| EcomQ2IRetrieval | ID, MS, PT, TH, TL, VI, ZH | Given an ecommerce query, retrieve a relevant item |
| EcomQueryRewriteRetrieval | ID, MS, PT, TH, TL, VI, ZH | Given an ecommerce query, retrieve another query with a similar meaning |
| **Classification** | | |
| EcomItemL1Classification | ID, MS, PT, TH, TL, VI, ZH | RGiven an e-commerce item title, predict its L1 product category |
| EcomItemL2Classification | ID, MS, PT, TH, TL, VI, ZH | Given an e-commerce item title, predict its L2 product category |

Table 14 | E-Commerce Evaluation Datasets